\documentclass[10pt,twocolumn,letterpaper]{article}

\usepackage[pagenumbers]{cvpr} 
\makeatletter
\@namedef{ver@everyshi.sty}{}
\makeatother

\usepackage[accsupp]{axessibility}  
\usepackage{chittesh}
\usepackage{graphicx}
\usepackage{amsmath}
\usepackage{amssymb}
\usepackage{booktabs}
\usepackage[utf8]{inputenc} 
\usepackage[T1]{fontenc}    
\usepackage{url}            
\usepackage{amsfonts}       
\usepackage{nicefrac}       
\usepackage{microtype}      
\usepackage{multirow}
\usepackage[table,xcdraw]{xcolor}
\usepackage{diagbox}
\usepackage{svg}
\usepackage{caption}
\usepackage{adjustbox}
\usepackage{mathtools}
\usepackage{algorithm}
\usepackage[noend]{algpseudocode}
\usepackage{setspace}

\algnewcommand{\LineComment}[1]{\State \(\triangleright\) #1}

\usepackage[pagebackref,breaklinks,colorlinks]{hyperref}

\usepackage[capitalize]{cleveref}
\crefname{section}{Sec.}{Secs.}
\Crefname{section}{Section}{Sections}
\Crefname{table}{Table}{Tables}
\crefname{table}{Tab.}{Tabs.}

\def\cvprPaperID{4750}
\def\confName{CVPR}
\def\confYear{2023}

\begin{document}


\newif\ifstandalonesupplement



\ifstandalonesupplement
    \title{Supplementary Material for Learning to Zoom and Unzoom}
\else
    \title{Learning to Zoom and Unzoom}
\fi

\setcounter{footnote}{1}

\author{Chittesh Thavamani\textsuperscript{1}\qquad Mengtian Li\thanks{Now at Waymo.}\textsuperscript{\enspace 1}\qquad Francesco Ferroni\thanks{Now at Nvidia.}\textsuperscript{\enspace 2}\qquad Deva Ramanan\textsuperscript{1}\qquad\\
\textsuperscript{1}Carnegie Mellon University\qquad \textsuperscript{2}Argo AI\\
{\tt\small tchittesh@gmail.org\qquad mengtial@alumni.cmu.edu\qquad fferroni@nvidia.com\qquad deva@cs.cmu.edu}
}
\maketitle

\ifstandalonesupplement
    \input{standalone_appendix}
    {\small
    \bibliographystyle{ieee_fullname}
    \bibliography{egbib}
    }
\else
    
\begin{abstract}


Many perception systems in mobile computing, autonomous navigation, and AR/VR face strict compute constraints that are particularly challenging for high-resolution input images.
Previous works propose nonuniform downsamplers that "learn to zoom" on salient image regions, reducing compute while retaining task-relevant image information.
However, for tasks with spatial labels (such as 2D/3D object detection and semantic segmentation), such distortions may harm performance.
In this work (LZU), we "learn to zoom" in on the input image, compute spatial features, and then "unzoom" to revert any deformations.
To enable efficient and differentiable unzooming, we approximate the zooming warp with a piecewise bilinear mapping that is invertible.
LZU can be applied to any task with 2D spatial input and any model with 2D spatial features, and we demonstrate this versatility by evaluating on a variety of tasks and datasets: \emph{object detection} on Argoverse-HD, \emph{semantic segmentation} on Cityscapes, and \emph{monocular 3D object detection} on nuScenes.
Interestingly, we observe boosts in performance even when high-resolution sensor data is unavailable, implying that LZU can be used to "learn to upsample" as well.
Code and additional visuals are available at \url{https://tchittesh.github.io/lzu/}.

\end{abstract}

\section{Introduction}

In many applications, the performance of perception systems is bottlenecked by strict inference-time constraints.
This can be due to limited compute (as in mobile computing), a need for strong real-time performance (as in autonomous vehicles), or both (as in augmented/virtual reality).
These constraints are particularly crippling for settings with high-resolution sensor data.
Even with optimizations like model compression~\cite{cheng2017survey} and quantization~\cite{37631}, it is common practice to downsample inputs during inference. 

However, running inference at a lower resolution undeniably destroys information. 
While some information loss is unavoidable, the usual solution of uniform downsampling assumes that each pixel is equally informative towards the task at hand.
To rectify this assumption, Recasens~\etal~\cite{recasens2018learning} propose Learning to Zoom (LZ), a nonuniform downsampler that samples more densely at salient (task-relevant) image regions.
They demonstrate superior performance relative to uniform downsampling on human gaze estimation and fine-grained image classification. However, this formulation warps the input image and thus requires labels to be invariant to such deformations.

\begin{figure}
\centering
\includegraphics[width=\linewidth]{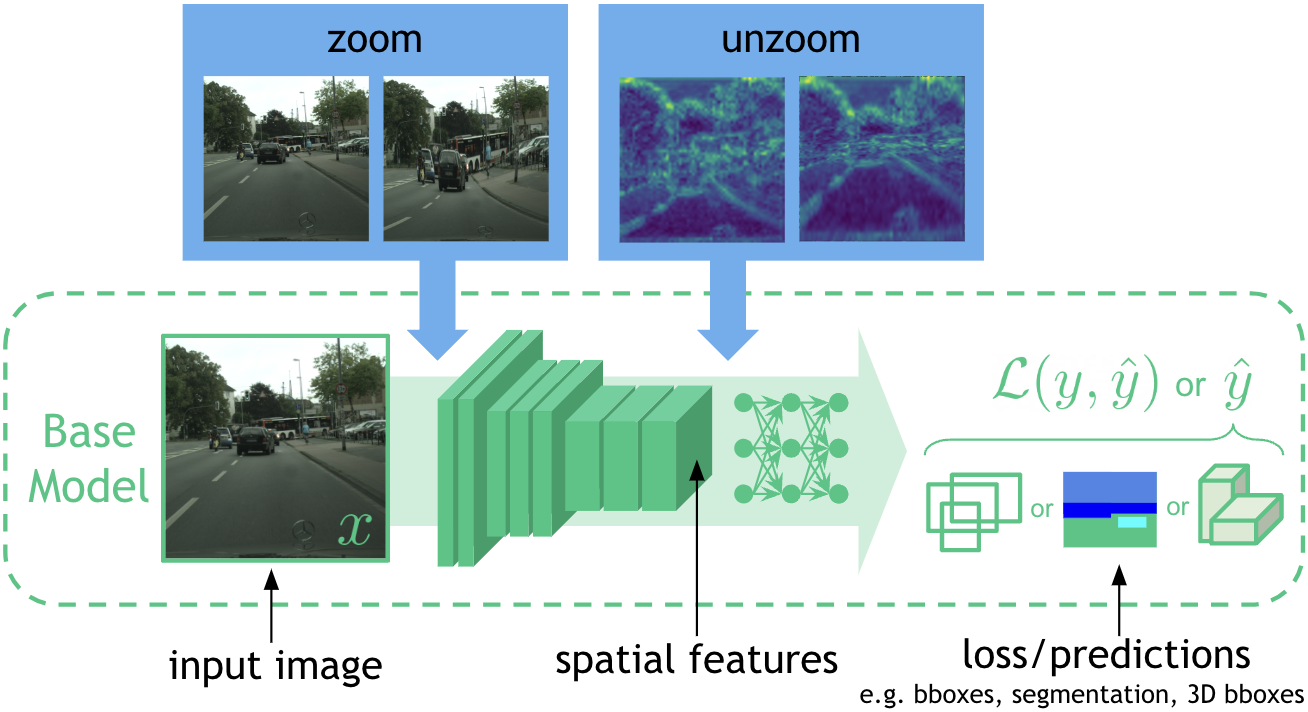}
\caption{LZU is characterized by "zooming" the input image, computing spatial features, then "unzooming" to revert spatial deformations. LZU can be applied to any task and model that makes use of internal 2D features to process 2D inputs. We show visual examples of output tasks including 2D detection, semantic segmentation, and 3D detection from RGB images.}
\label{fig:teaser}
\end{figure}

Adapting LZ downsampling to tasks with spatial labels is trickier, but has been accomplished in followup works for semantic segmentation (LDS~\cite{jin2021learning}) and 2D object detection (FOVEA~\cite{thavamani2021fovea}). LDS~\cite{jin2021learning} does not unzoom during learning, and so defines losses in the warped space. This necessitates additional regularization that may not apply to non-pixel-dense tasks like detection. FOVEA~\cite{thavamani2021fovea} {\em does} unzoom bounding boxes for 2D detection, but uses a special purpose solution that avoids computing an inverse, making it inapplicable to pixel-dense tasks like semantic segmentation.
Despite these otherwise elegant solutions, there doesn't seem to be a general task-agnostic solution for intelligent downsampling.

Our primary contribution is a general framework in which we zoom in on an input image, process the zoomed image, and then {\em un}zoom the output back with an inverse warp. Learning to Zoom and Unzoom (LZU) can be applied to {\em any} network that uses 2D spatial features to process 2D spatial inputs (Figure~\ref{fig:teaser}) {\em with no adjustments to the network or loss}. To unzoom, we approximate the zooming warp with a piecewise bilinear mapping. This allows efficient and differentiable computation of the forward and inverse warps.

To demonstrate the generality of LZU, we demonstrate performance a variety of tasks: \emph{object detection} with RetinaNet~\cite{lin2017focal} on Argoverse-HD~\cite{li2020towards}, \emph{semantic segmentation} with PSPNet\cite{zhao2017pyramid} on Cityscapes~\cite{cordts2016cityscapes}, and \emph{monocular 3D detection} with FCOS3D~\cite{wang2021fcos3d} on nuScenes~\cite{nuScenes}. In our experiments, to maintain favorable accuracy-latency tradeoffs, we use cheap sources of saliency (as in~\cite{thavamani2021fovea}) when determining where to zoom. On each task, LZU increases performance over uniform downsampling and prior works with minimal additional latency.

Interestingly, for both 2D and 3D object detection, we also see performance boosts even when processing low resolution input data. While prior works focus on performance improvements via intelligent downsampling~\cite{recasens2018learning,thavamani2021fovea}, our results show that LZU can also improve performance by intelligently {\em up}sampling (suggesting that current networks struggle to remain scale invariant for small objects, a well-known observation in the detection community~\cite{lin2014microsoft}).

\section{Related Work}

We split related work into two sections. The first discusses the broad class of methods aiming to improve efficiency by paying "attention" to specific image regions. The second delves into works like LZU that accomplish this by differentiably resampling the input image.

\subsection{Spatial Attentional Processing}

By design, convolutional neural networks pay equal "attention" (perform the same computations) to all regions of the image. In many cases, this is suboptimal, and much work has gone into developing attentional methods that resolve this inefficiency.

One such method is Dynamic Convolutions~\cite{verelst2020dynamic}, which uses sparse convolutions to selectively compute outputs at only the salient regions. 
Similarly, gated convolutions are used in \cite{kong2019pixel, xie2020spatially}.
Notably, these methods implement "hard" attention in that the saliency is binary, and non-salient regions are ignored completely. 

Deformable Convolutions~\cite{dai2017deformable, zhu2019deformable} provides a softer implementation of spatial attention by learning per pixel offsets when applying convolutions, allowing each output pixel to attend adaptively to pixels in the input image. SegBlocks~\cite{verelst2020segblocks} also provides a softer attention mechanism by splitting the image into blocks and training a lightweight reinforcement learning policy to determine whether each block should be processed at a high or low resolution. This is akin to our method, which also has variable resolution, albeit in a more continuous manner. Our method is also generalizable to tasks in which it's infeasible to "stitch" together outputs from different blocks of the image (e.g. in detection where an object can span multiple blocks).

\subsection{Spatial Attention via Differentiable Image Resampling}

Spatial Transformer Networks~\cite{jaderberg2015spatial} introduces a differentiable method to resample an image. They originally propose this to invert changes in appearance due to viewpoint, thereby enforcing better pose invariance.

Learning to Zoom (LZ)~\cite{recasens2018learning} later adapts this resampling operation to "zoom" on salient image regions, acting as a spatial attention mechanism. Their key contribution is a transformation parameterized by a saliency map such that regions with higher saliency are more densely sampled. However, this deforms the image, requiring the task to have non-spatial labels.

Followup works~\cite{marin2019efficient, thavamani2021fovea, jin2021learning} adapt LZ downsampling to detection and semantic segmentation. For object detection, FOVEA~\cite{thavamani2021fovea} exploits the fact that image resampling is implemented via an inverse mapping to map predicted bounding boxes back into the original image space. This allows all processing to be done in the downsampled space and the final bounding box regression loss to be computed in the original space. However, when there are intermediate losses, as is the case with two-stage detectors containing region proposal networks (RPNs)~\cite{ren2015faster}, this requires more complex modifications to the usual delta loss formulation, due to the irreversibility of the inverse mapping.
For semantic segmentation, Jin~\etal~\cite{jin2021learning} apply LZ downsampling to both the input image and the ground truth and computes the loss in the downsampled space. This is elegant and model-agnostic but leads to misalignment between the training objective and the desired evaluation metric. In the extreme case, the model learns degenerate warps that sample "easy" parts of the image to reduce the training loss. To address this, they introduce additional regularization on the downsampler. Independently,~\cite{marin2019efficient} handcraft an energy minimization formulation to sample more densely at semantic boundaries. 

In terms of warping and unwarping, the closest approach to ours is Dense Transformer Networks~\cite{li2017dense}, which also inverts deformations introduced by nonuniform resampling. However, their warping formulation is not saliency-based, which makes it hard to work with spatial or temporal priors and also makes it time-consuming to produce the warping parameters.
Additionally, they only show results for semantic segmentation, whereas we show that our formulation generalizes across spatial vision tasks.

\section{Background}

Since our method is a generalization of previous works~\cite{recasens2018learning,thavamani2021fovea,jin2021learning}, we include this section as a condensed explanation of prerequisite formulations critical to understanding LZU.

\subsection{Image Resampling}

Suppose we want to resample an input image $\bI(\bx)$ to produce an output image $\bI'(\bx)$, both indexed by spatial coordinates $\bx \in [0,1]^2$. 
Resampling is typically implemented via an {\em inverse} map $\calT: [0,1]^2 \to [0,1]^2$ from output to input coordinates~\cite{beier1992feature}.
For each output coordinate, the inverse map computes the source location from which to "steal" the pixel value, i.e. $\bI'(\bx) = \bI(\calT(\bx))$.
In practice, we are often given a discretized input image $\bI \in \R^{H \times W \times C}$ and are interested in computing a discretized output $\bI' \in \R^{H' \times W' \times C}$. 
To do so, we compute $\bI'(\bx)$ at grid points $\bx \in \mathrm{Grid}(H',W')$, where $\mathrm{Grid}(H,W) := \mathrm{Grid}(H) \times \mathrm{Grid}(W)$ and $\mathrm{Grid}(D) := \{\frac{d-1}{D-1} : d \in [D]\}$. However, $\calT(\bx)$ may return non-integer pixel locations at which the exact value of $\bI$ is unknown. In such cases, we use bilinear interpolation to compute $\bI(\calT(\bx))$. As proven in~\cite{jaderberg2015spatial}, such image resampling is differentiable with respect to $\calT$ and $\bI$. 

\subsection{Saliency-Guided Downsampling}
\label{sec:saliency-downsampling}

\begin{figure}[t]
\centering
\includegraphics[width=\columnwidth]{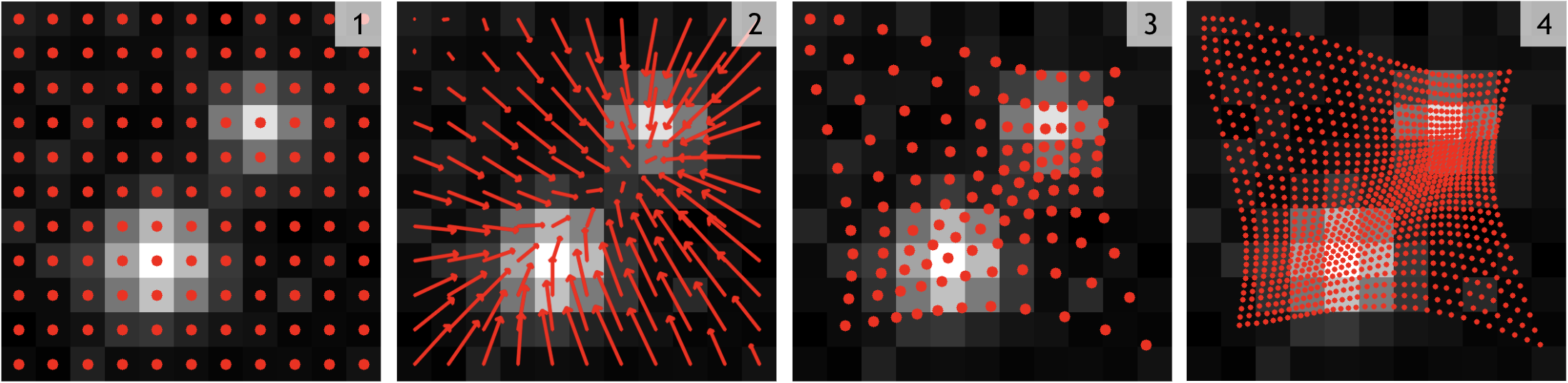}
\caption{
Illustration of $\calT_\mathrm{LZ}$~\cite{recasens2018learning}. Suppose we have a saliency map $\bS \in \R^{h\times w}$ (visualized in the background) and want a warped image of size $H' \times W'$. (1) We start with a uniform grid of sample locations $\mathrm{Grid}(h, w)$. (2) Grid points are "attracted" to nearby areas with high saliency. (3) Applying this "force" yields $\calT_\mathrm{LZ}[\mathrm{Grid}(h,w)]$. (4) Bilinear upsampling yields $\widetilde\calT_\mathrm{LZ}[\mathrm{Grid}(H', W')]$.
}
\label{fig:LZ_downsampling}
\end{figure}

\begin{figure}[t]
\centering
\includegraphics[width=\columnwidth]{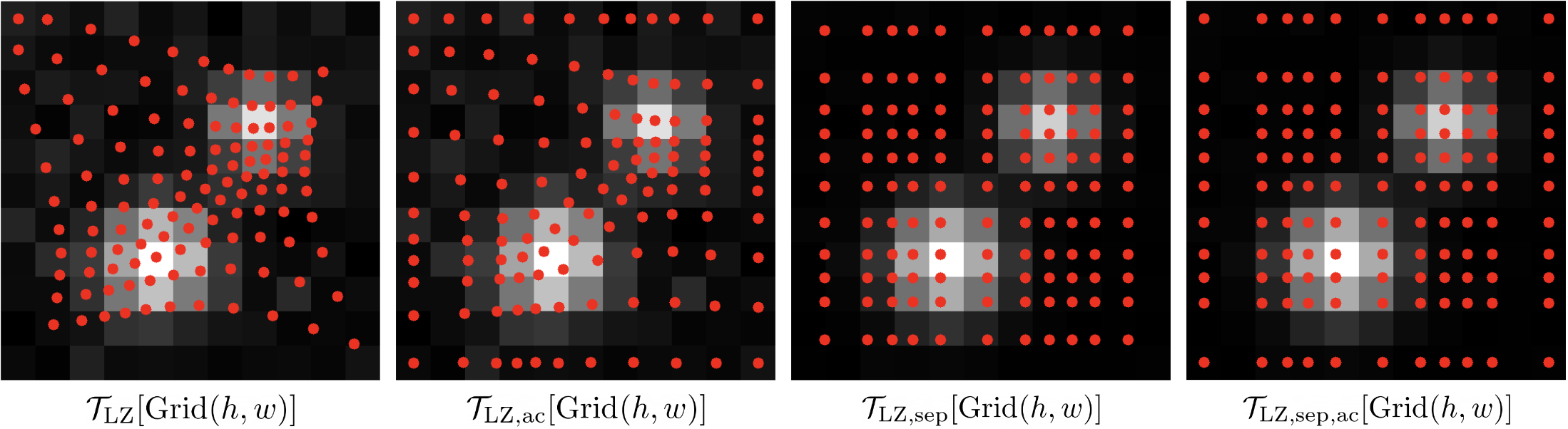}
\caption{
Examples of the anti-cropping ($\mathrm{ac}$) and separable ($\mathrm{sep}$) variants of $\calT_\mathrm{LZ}$ from~\cite{thavamani2021fovea}.
}
\label{fig:LZ_downsampling_variants}
\end{figure}

When using nonuniform downsampling for information retention, it is useful to parameterize $\calT$ with a saliency map $\bS(\bx)$ representing the desired sample rate at each spatial location $\bx \in [0,1]^2$~\cite{recasens2018learning}.
Recasens~\etal~\cite{recasens2018learning} go on to approximate this behavior by having each sample coordinate $\calT(\bx)$ be "attracted" to nearby areas $\bx'$ with high saliency $\bS(\bx')$ downweighted according to a distance kernel $k(\bx,\bx')$, as illustrated in Figure~\ref{fig:LZ_downsampling}. Concretely, $\calT_{\mathrm{LZ}}(\bx) = (\calT_{\mathrm{LZ}, x}(\bx), \calT_{\mathrm{LZ}, y}(\bx))$, where
\begin{equation}
    \calT_{\mathrm{LZ}, x}(\bx) =
    \frac{\int_{\bx'} \bS(\bx') k(\bx, \bx') \bx'_x\,d\bx'}{\int_{\bx'} \bS(\bx') k(\bx, \bx') \,d\bx'},
\end{equation}
\begin{equation}
    \calT_{\mathrm{LZ}, y}(\bx) =
    \frac{\int_{\bx'} \bS(\bx') k(\bx, \bx') \bx'_y \,d\bx'}{\int_{\bx'} \bS(\bx') k(\bx, \bx') \,d\bx'}.
\end{equation}

\cite{thavamani2021fovea} proposes \emph{anti-cropping} and \emph{separable} variants of this downsampler. The anti-cropping variant $\calT_{\mathrm{LZ,ac}}$ prevents the resampling operation from cropping the image. The separable variant marginalizes the saliency map $\bS(\bx)$ into two 1D saliency maps $\bS_x(x)$ and $\bS_y(y)$, and replaces the kernel $k(\bx, \bx')$ with a two 1D kernels $k_x$ and $k_y$ (although generally $k_x = k_y$).
Then, $\calT_{\mathrm{LZ,sep}}(\bx) = \left(\calT_{\mathrm{LZ,sep,x}}(\bx_x), \calT_{\mathrm{LZ,sep,y}}(\bx_y)\right)$ where
\begin{equation}
\calT_{\mathrm{LZ,sep,x}}(x) = 
\frac{\int_{x'} \bS_x(x')k_x(x, x')x'\,dx'}{\int_{x'} \bS_x(x')k_x(x, x')\,dx'},
\end{equation}
\begin{equation}
\calT_{\mathrm{LZ,sep,y}}(y) = 
\frac{\int_{y'} \bS_y(y')k_y(y, y')y'\,dy'}{\int_{y'} \bS_y(y')k_y(y, y')\,dy'}.
\end{equation}
This preserves axis-alignment of rectangles, which is crucial to object detection where bounding boxes are specified via corners. We refer to the above method and all variants as \emph{LZ downsamplers}, after the pioneering work "Learning to Zoom"~\cite{recasens2018learning}. Examples of each variant are shown in Figure~\ref{fig:LZ_downsampling_variants}.






\section{Method}

We begin by discussing our general technique for warp inversion. Then, we discuss the LZU framework and how we apply warp inversion to efficiently "unzoom".

\subsection{Efficient, Differentiable Warp Inversion}
\label{sec:warp-inv}

\begin{figure}
    \centering
    \includegraphics[width=\columnwidth]{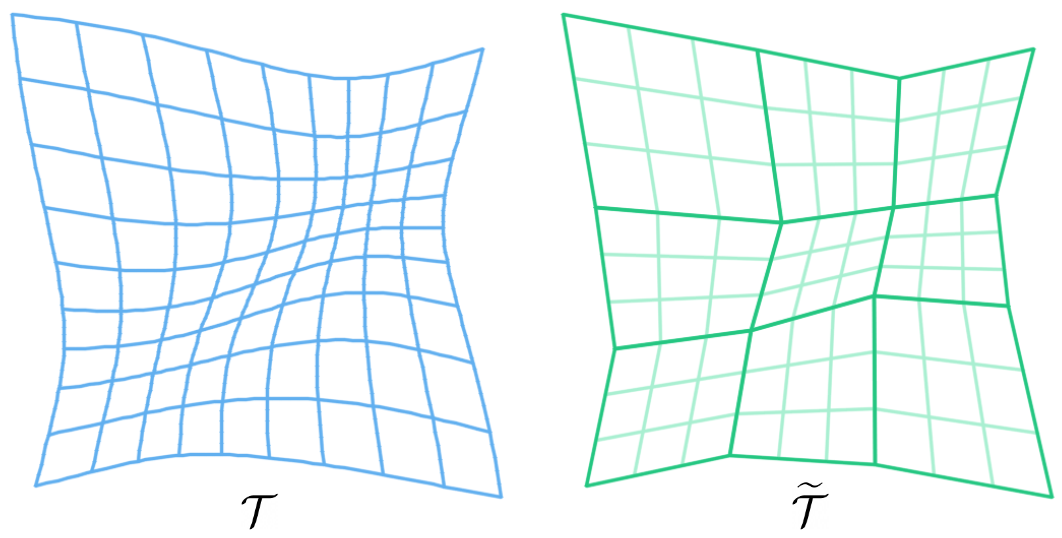}
    \caption{Given a warp $\calT$, we construct an approximation $\widetilde\calT$ designed for efficient inversion. As illustrated, $\widetilde\calT$ is a piecewise tiling of simpler invertible maps. This allows us to approximate the inverse $\widetilde\calT^{-1}$, even when $\calT^{-1}$ lacks a closed form.}
    \label{fig:warp_approximation}
\end{figure}

Suppose we have a continuous map $\calT: [0,1]^2 \to [0,1]^2$.
Our primary technical innovation is an efficient and differentiable approximation of $\calT^{-1}$, even in cases where $\calT$ has no closed-form inverse.

Since $\calT$ is potentially difficult to invert, we first approximate it as $\widetilde\calT$, a piecewise tiling of simpler invertible transforms (illustrated in Figure~\ref{fig:warp_approximation}). Formally,
\begin{equation}
    \widetilde\calT = \bigcup_{\substack{i \in [h-1]\\j \in [w-1]}} \widetilde\calT_{ij},
\end{equation}
where the $ij$-th tile $\widetilde\calT_{ij}$ is any bijective map from the rectangle formed by corners $R_{ij}=\{\frac{i-1}{h-1}, \frac{i}{h-1}\} \times \{\frac{j-1}{w-1}, \frac{j}{w-1}\}$ to quadrilateral $\calT[R_{ij}]$. For our purposes, we choose bilinear maps as our tile function, although homographies could work just as well. Then, so long as $\widetilde\calT$ is injective (if each of the tiles $\widetilde\calT_{ij}$ is nondegenerate and no two tiles overlap), we are guaranteed a well-defined left inverse $\widetilde\calT^{-1}: [0,1]^2 \to [0,1]^2$ given by
\begin{equation}
    \widetilde\calT^{-1}(\bx) = 
    \begin{cases}
    \widetilde\calT^{-1}_{ij}(\bx) & \text{if }\bx \in \mathrm{Range}(\widetilde\calT_{ij}) \\
    0 & \text{else}
    \end{cases}. \label{eq:inf}
\end{equation}

Equation \ref{eq:inf} is {\bf efficient} to compute, since determining if $\bx \in \mathrm{Range}(\widetilde\calT_{ij})$ simply involves checking if $\bx$ is in the quadrilateral $\calT[R_{ij}]$ and computing the inverse $\widetilde\calT^{-1}_{ij}$ of a bilinear map amounts to solving a quadratic equation~\cite{wolberg1990digital}.
This efficiency is crucial to maintaining favorable accuracy-latency tradeoffs. $\widetilde\calT^{-1}$ is guaranteed to be {\bf differentiable} with respect to $\calT$, since for each $\bx \in \widetilde\calT[R(i,j)]$, the inverse bilinear map can be written as a quadratic function of the four corners of tile $ij$ (see Appendix~\ref{sec:bilinear-maps} for exact expression).
This allows gradients to flow back into $\calT$, letting us learn the parameters of the warp.

In the case of LZ warps, $\calT_{\mathrm{LZ}}$ has no closed form inverse to the best of our knowledge. 
Because $\calT_{\mathrm{LZ}}[\mathrm{Grid}(h, w)]$ has no foldovers~\cite{recasens2018learning}, $\widetilde\calT_{\mathrm{LZ}}$ must be injective, implying its inverse $\widetilde\calT_{\mathrm{LZ}}^{-1}$ is well-defined.

When applying an LZ warp, saliency can be learned (with trainable parameters) or unlearned (with fixed parameters), and fixed (invariant across frames) or adaptive (changes every frame).
Adaptive saliency maps require efficient warp inversion since a different warp must be applied on every input. Learned saliency maps require differentiability. We note that fixed unlearned saliency maps do not technically require efficiency or differentiability, and most of our current results show that such saliency maps are already quite effective, outperforming prior work. We posit that LZU would shine even more in the learned adaptive setting, where it could make use of temporal priming for top-down saliency.

\subsection{Learning to Zoom and Unzoom}
\label{sec:lzu}

In the Learning to Zoom and Unzoom (LZU) framework, we use existing LZ downsamplers (see Section~\ref{sec:saliency-downsampling}) to "zoom" in on the input image, compute spatial features, and then use our warp inversion formulation to "unzoom" and revert any deformations in the feature map, as shown in Figure~\ref{fig:teaser}. This framework is applicable to all tasks with 2D spatial input and all models with some intermediate 2D spatial representation.

Notice that a poorly approximated inverse warp $\widetilde\calT^{-1}$ would lead to misaligned features and a drop in performance.
As a result, we use the approximate forward warp $\widetilde\calT$ instead of the true forward warp $\calT$, so that the composition of forward and inverse warps is \textit{actually} the identity function.
See Appendix~\ref{sec:approximations} for a discussion of the associated tradeoff.

\begin{figure}[t]
    \centering
    \includegraphics[width=\linewidth]{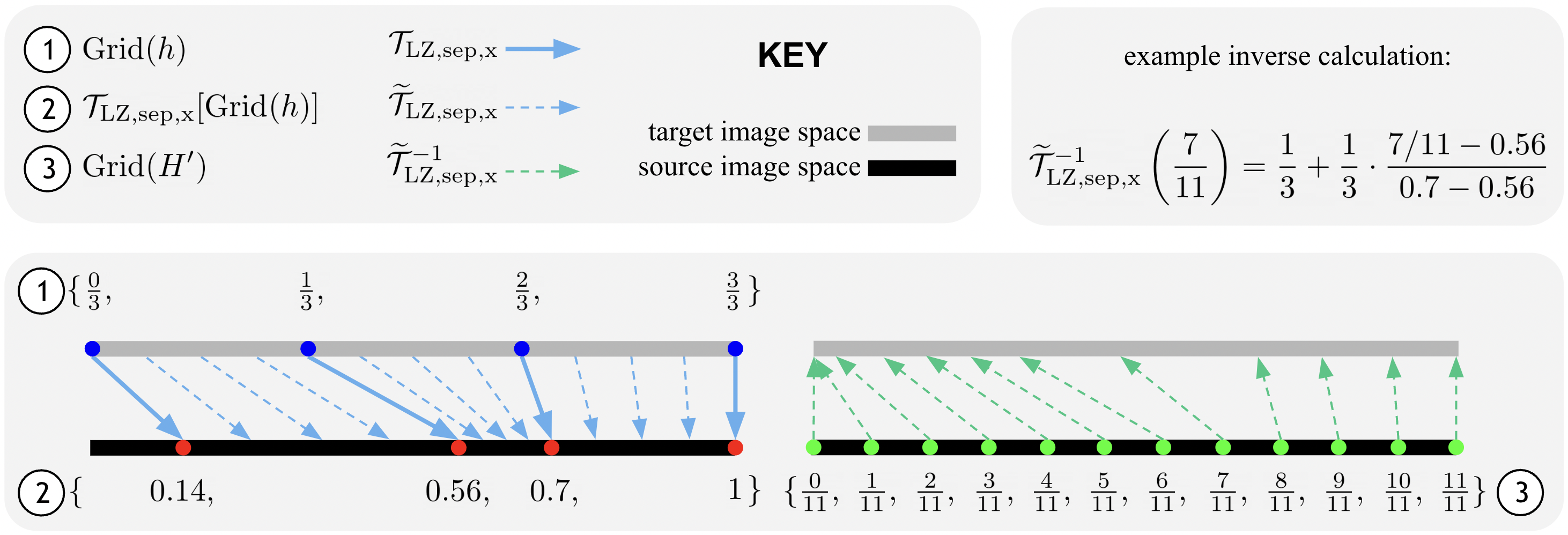}
    \caption{Inverting each axis of a separable warp. 
    LZU first evaluates the forward warp $\calT_\mathrm{LZ,sep,x}$ (solid blue arrows) at a uniform grid of target locations (blue points).
    The resulting source locations are shown as red points.
    LZU then approximates the warp in between these samples via a \textit{linear} transform; this piecewise linear map is $\widetilde\calT_\mathrm{LZ,sep,x}$ (dotted blue arrows).
    To evaluate the inverse $\widetilde\calT^{-1}_\mathrm{LZ,sep,x}$ (dotted green arrows), we must determine for each green point which red points it falls between and invert the corresponding linear transform. An example is shown in the top-right. 
    }
    \label{fig:unzoom-axis}
\end{figure}

To maintain favorable accuracy-latency tradeoffs, we make several optimizations to our forward and inverse warps. As done in previous works~\cite{jin2021learning,recasens2018learning,thavamani2021fovea}, for the forward warp or "zoom," instead of computing $\calT_{\mathrm{LZ}}[\mathrm{Grid}(H', W')]$, we compute $\calT_{\mathrm{LZ}}[\mathrm{Grid}(h, w)]$ for smaller $h \ll H'$ and $w \ll W'$ and bilinearly upsample this to get $\widetilde\calT_{\mathrm{LZ}}[\mathrm{Grid}(H', W')]$. This also reduces the complexity of computing the inverse, by reducing the number of cases in our piecewise bilinear map from $H'\cdot W'$ to $h\cdot w$.

We explore efficient implementations of both separable and nonseparable warp inversion, but we find experimentally that nonseparable warps perform no better than separable warps for a strictly higher latency cost, so we use separable warps for our experiments. Details for efficiently inverting nonseparable warps are given in Appendix~\ref{sec:nonsep-inversion}.
For separable warps $\calT_{\mathrm{LZ,sep}}$, we invert each axis separately and take the Cartesian Product:
\begin{align}
\widetilde\calT_{\mathrm{LZ,sep}}^{-1}[\mathrm{Grid}&(H', W')]
= \\
&\widetilde\calT_{\mathrm{LZ,sep,x}}^{-1}[\mathrm{Grid}(H')]
\times
\widetilde\calT_{\mathrm{LZ,sep,y}}^{-1}[\mathrm{Grid}(W')]. \nonumber
\end{align}
This further reduces our problem from inverting a piecewise bilinear map with $h \cdot w$ pieces to inverting two piecewise \emph{linear} maps with $h$ and $w$ pieces each. Figure~\ref{fig:unzoom-axis} visualizes how to invert each axis.

When unwarping after feature pyramid networks (FPNs)~\cite{lin2017feature}, we may have to evaluate the inverse $\widetilde\calT_{\mathrm{LZ}}^{-1}$ at multiple resolutions $\mathrm{Grid}(H', W')$, $\mathrm{Grid}(H' / 2, W' / 2)$, etc. In practice, we evaluate $\widetilde\calT_{\mathrm{LZ}}^{-1}[\mathrm{Grid}(H', W')]$ and then approximate the inverse at lower resolutions via bilinear downsampling.
This is surprisingly effective (see Appendix~\ref{sec:approximations}) and leads to no observable loss in performance.

Finally, as introduced in~\cite{thavamani2021fovea}, we can also use a fixed warp to exploit dataset-wide spatial priors, such as how objects are concentrated around the horizon in many autonomous driving datasets. This allows us to cache forward and inverse warps, greatly reducing additional latency.




\section{Experiments}
\label{sec:exp}

First, we compare LZU to naive uniform downsampling and previous works on the tasks of 2D object detection and semantic segmentation. We include ablations to evaluate the effectiveness of training techniques and explore the upsampling regime. Then, we evaluate LZU on monocular 3D object detection, a task which no previous works have applied "zooming" to. We perform all timing experiments with a batch size of 1 on a single RTX 2080 Ti GPU. Figure~\ref{fig:qualitative-results} contains qualitative results and analysis across all tasks. Full implementation details and hyperparameters are given in Appendix~\ref{sec:impl-details}.

\subsection{2D Object Detection}
\label{sec:2ddet}

\begin{table}[t]
\centering
\footnotesize
\setlength{\tabcolsep}{2pt} 
\resizebox{\columnwidth}{!}{%
\begin{tabular}{rlccccccc}
\toprule
Scale & Method  & AP & AP$_{50}$ & AP$_{75}$ & AP$_{S}$ & AP$_{M}$ & AP$_{L}$ & Lat (ms) \\
\midrule
0.25x & Uniform & 10.5&	18.0&	9.9	&0.3&	5.2	&38.6&\textbf{23.3}\\
0.25x & LZU, fixed & \textbf{12.4}	&22.6&	11.2&	1.0	&\textbf{7.1}&	\textbf{39.2}&23.6\\
0.25x & LZU, adaptive &12.3	&\textbf{22.8}&	\textbf{11.3}&	\textbf{1.4}&	6.6	&38.0 & 26.4 \\
\midrule
0.5x  & Uniform & 22.6	& 38.7	& 21.7	& 3.7	& 22.1	& \textbf{53.1} & \textbf{36.0} \\
0.5x  & FOVEA~\cite{thavamani2021fovea} & 24.9	& 40.3	& \textbf{25.3}	& \textbf{7.1}	& \textbf{27.7}	& 50.6 & 37.9 \\
0.5x & LZU, fixed & 25.2	& 42.1&	24.8&	5.5	&26.7&	51.8 & 36.4 \\
0.5x  & LZU, adaptive & \textbf{25.3}	& \textbf{43.0} &	24.6 &	6.1 &	25.9 &	52.6 & 39.3 \\
0.5x  & LZU, adaptive & 22.8&	39.3&	22.3&	5.1	&22.7&	48.9 & 39.3 \\
&  w/o cascade sal. \\
\midrule
0.75x  & Uniform & 29.5	& 48.4 &	29.6&	9.1&	32.4&	\textbf{55.1} & \textbf{62.9} \\
0.75x & LZU, fixed & \textbf{30.8}	&\textbf{50.4}&	\textbf{31.8}&	\textbf{10.9}&	\textbf{33.5}	&54.1	&63.5 \\
0.75x & LZU, adaptive & 26.5&	44.6&	26.7&	8.3	&28.7&	48.7 &66.3 \\
\midrule
1x & Uniform & 31.9&	51.5&	33.1&	11.4&	35.9&	54.5&	\textbf{98.3}\\
1x & LZU, fixed & \textbf{32.6}&	\textbf{52.8}&	\textbf{34.0}&	\textbf{13.2}&	36.0	&\textbf{54.7}	&99.3\\
1x & LZU, adaptive & 32.0&	52.4&	33.1&	12.5&	\textbf{36.3}&	52.9	& 102.0 \\
\bottomrule
\end{tabular}
}
\caption{2D object detection results of RetinaNet~\cite{lin2017focal} on Argoverse-HD~\cite{li2020towards}.
Fixed LZU uses a dataset-wide spatial prior, and adaptive LZU uses a temporal prior based on previous frame detections.
LZU consistently outperforms the uniform downsampling baseline and prior work across all scales, with additional latency less than $4$ms.
We hypothesize that the drop in AP$_L$ is because objects that are already large benefit less from zooming.
Still, this drawback is offset by larger improvements on small and medium objects.
}
\label{tab:det-avhd}
\end{table}

\begin{table}[t]
\setlength{\tabcolsep}{3pt} 
\footnotesize
\centering
2D Object Detection
\vspace{0.5em}
\begin{tabular}{rccccc@{\hskip 0.3cm}rcccc}
    \toprule
    \multicolumn{5}{c}{Uniform Resampling} & & \multicolumn{5}{c}{LZU Resampling} \\
    \cmidrule(){1-5} \cmidrule(){7-11}
    & \multicolumn{4}{c}{From} & & &  \multicolumn{4}{c}{From} \\
    \cmidrule(r){2-5}\cmidrule(){8-11}
    To~~ & 0.25x & 0.5x & 0.75x & 1x & & To~~ & 0.25x & 0.5x & 0.75x & 1x \\
    \cmidrule(r){1-1}\cmidrule(r){2-5}\cmidrule(r){7-7}\cmidrule(){8-11}
    0.25x & 10.5 &	\textcolor{blue}{10.5} &	\textcolor{blue}{10.5}&	\textcolor{blue}{10.5}
    & & 0.25x & {\textbf{11.7}}&	\textcolor{blue}{\textbf{12.4}} &	\textcolor{blue}{\textbf{12.4}}&	\textcolor{blue}{\textbf{12.4}}\\
    0.5x &\textcolor{red}{17.0}& 22.6 &	\textcolor{blue}{22.6}&	\textcolor{blue}{22.6} & & 0.5x &\textcolor{red}{\textbf{20.9}}&	\textbf{24.8}&	\textcolor{blue}{\textbf{24.8}}&	\textcolor{blue}{\textbf{25.2}}\\
    0.75x &\textcolor{red}{\textbf{23.5}}&	\textcolor{red}{28.5}&	29.5 &	\textcolor{blue}{29.5} & & 0.75x &\textcolor{red}{22.5}&	\textcolor{red}{\textbf{29.4}}&	{\textbf{30.0}}&	\textcolor{blue}{\textbf{30.8}}\\
    1x& \textcolor{red}{13.5}	&\textcolor{red}{28.4}&	\textcolor{red}{30.9}&	31.9 & & 1x& \textcolor{red}{\textbf{22.1}}	&\textcolor{red}{\textbf{30.7}}&	\textcolor{red}{\textbf{31.2}}&	{\textbf{32.6}} \\
    \bottomrule
\end{tabular}
Monocular 3D Object Detection
\begin{tabular}{rccccc@{\hskip 0.3cm}rcccc}
    \toprule
    \multicolumn{5}{c}{Uniform Resampling} & & \multicolumn{5}{c}{LZU Resampling} \\
    \cmidrule(){1-5} \cmidrule(){7-11}
    & \multicolumn{4}{c}{From} & & &  \multicolumn{4}{c}{From} \\
    \cmidrule(r){2-5}\cmidrule(){8-11}
    To~~ & 0.25x & 0.5x & 0.75x & 1x & & To~~ & 0.25x & 0.5x & 0.75x & 1x \\
    \cmidrule(r){1-1}\cmidrule(r){2-5}\cmidrule(r){7-7}\cmidrule(){8-11}
    0.25x & 21.8 &	\textcolor{blue}{21.8} &	\textcolor{blue}{21.8}&	\textcolor{blue}{21.8}
    & & 0.25x & {\textbf{22.5}}&	\textcolor{blue}{\textbf{23.5}} &	\textcolor{blue}{\textbf{23.4}}&	\textcolor{blue}{\textbf{23.4}}\\
    0.5x &\textcolor{red}{25.4}& 27.5 &	\textcolor{blue}{27.5}&	\textcolor{blue}{27.5} & & 0.5x &\textcolor{red}{\textbf{27.0}}&	{\textbf{29.2}}&	\textcolor{blue}{\textbf{29.1}}&	\textcolor{blue}{\textbf{29.3}}\\
    0.75x &\textcolor{red}{27.6}&	\textcolor{red}{30.3}&	30.5 &	\textcolor{blue}{30.5} & & 0.75x &\textcolor{red}{\textbf{29.0}}&	\textcolor{red}{\textbf{31.6}}&	{\textbf{31.6}}&	\textcolor{blue}{\textbf{31.8}}\\
    1x& \textcolor{red}{28.4}	&\textcolor{red}{30.7}&	\textcolor{red}{31.1} &	31.2 & & 1x& \textcolor{red}{\textbf{30.1}}	&\textcolor{red}{\textbf{32.5}}&	\textcolor{red}{\textbf{32.7}}&	{\textbf{32.6}} \\
    \bottomrule
\end{tabular}
\caption{2D and 3D object detection results in the \textcolor{red}{upsampling} and \textcolor{blue}{downsampling} regimes, using the "Uniform" and "LZU, fixed" models from Tables~\ref{tab:det-avhd} and~\ref{tab:3ddet}. LZU is surprisingly effective even in the upsampling regime! This demonstrates that simply allocating more pixels to small objects (without retaining extra information) can help performance, suggesting that detectors still struggle with scale invariance for small objects.}
\label{tab:upsampling}
\end{table}

\begin{figure*}[t]
\centering
\includegraphics[width=0.93\linewidth]{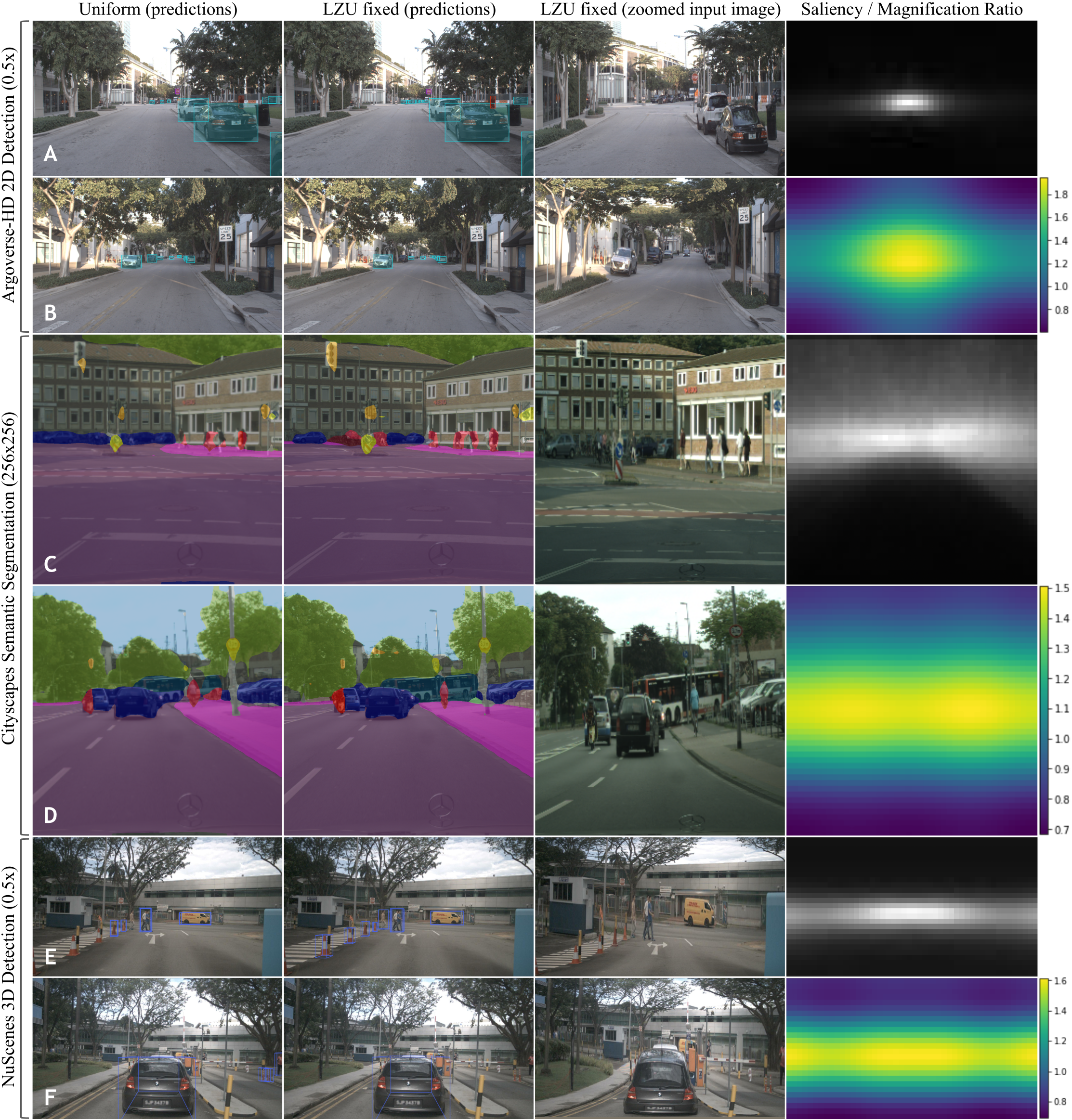}
\caption{
Examples of the success and failure cases of LZU.
Rows A and E show examples where zooming in on the horizon helps the detector pick up smaller objects.
On the other hand, sometimes zooming leads to false negatives, such as the black car in Row B and objects near the edge in Row F.
For segmentation, LZU consistently improves quality near the center of the image.
The last column shows the saliency map used in each case and the resulting spatial magnification ratios.
For the Argoverse-HD~\cite{li2020towards} dataset, the magnification ratio at the center is nearly $2$x, meaning the "zoom" is preserving nearly all information in that region, at the cost of information at the corners.
}
\label{fig:qualitative-results}
\end{figure*}

For 2D object detection, we evaluate LZU using RetinaNet~\cite{lin2017focal} (with a ResNet-50 backbone~\cite{he2016deep} and FPN~\cite{lin2017feature}) on Argoverse-HD~\cite{li2020towards}, an object detection dataset for autonomous driving with high resolution $1920 \times 1200$ videos.
For our baselines, we compare to uniform downsampling and FOVEA~\cite{thavamani2021fovea}, a previous work that applies LZ downsampling to detection by unwarping bounding boxes. We keep the same hyperparameters and setup as in FOVEA~\cite{thavamani2021fovea}.
Experiments are run at $0.25$x, $0.5$x, $0.75$x, and $1$x scales, to measure the accuracy-latency tradeoff. 

Our LZU models "unzoom" the feature map at each level after the FPN~\cite{lin2017feature}. We adopt the low-cost saliency generators introduced in~\cite{thavamani2021fovea} --- a "fixed" saliency map exploiting dataset-wide spatial priors, and an "adaptive" saliency map exploiting temporal priors by zooming in on detections from the previous frame. When training the adaptive version, we simulate previous detections by jittering the ground truth for the first two epochs. For the last epoch, we jitter \textit{detections} on the current frame to better simulate previous detections; we call this "cascaded" saliency. To determine saliency hyperparameters, we run grid search at $0.5$x scale on splits of the training set (details in Appendix~\ref{sec:sens-saliency},~\ref{sec:impl-details}). We use a learning rate of $0.01$ and keep all other training settings identical to the baseline. Latency is measured by timing only the additional operations (the "zoom" and "unzoom") and adding it to the baseline. This is done to mitigate the impact of variance in the latency of the backbone and detector head.

\begin{figure*}
\centering
\includegraphics[width=\linewidth]{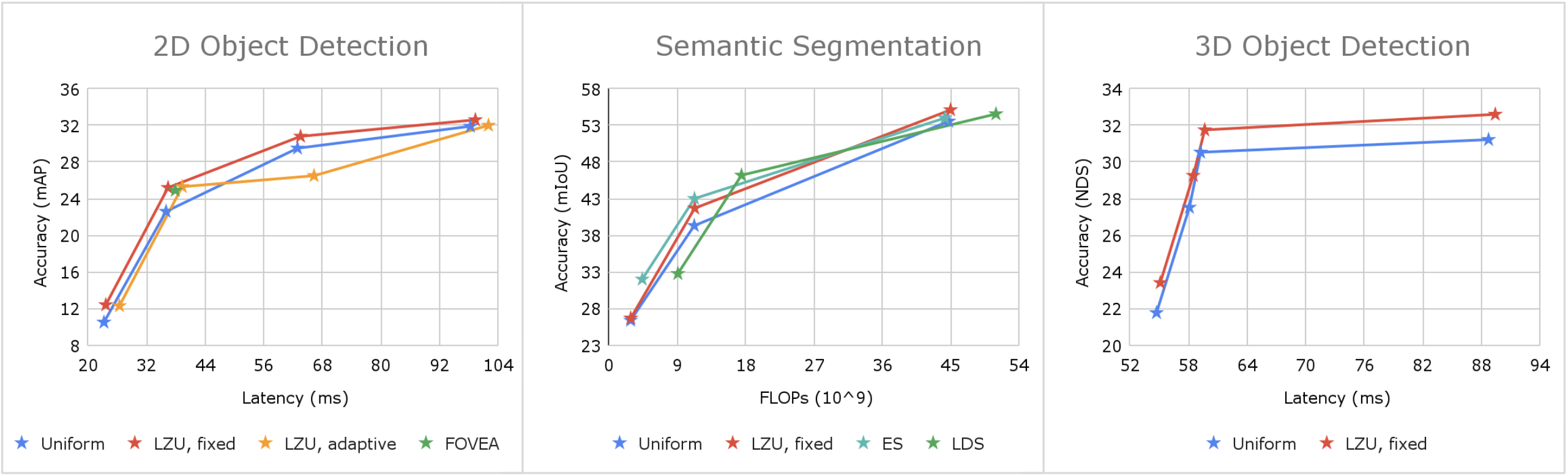}
\caption{
Plotting the accuracy-latency/FLOPs tradeoffs reveals the Pareto optimal methods for each task.
Fixed LZU is Pareto optimal for both 2D and 3D object detection, outperforming uniform downsampling and FOVEA~\cite{thavamani2021fovea}.
For semantic segmentation, we use FLOPs in lieu of latency to enable fair comparisons (ES~\cite{marin2019efficient} only reports FLOPS and LDS~\cite{jin2021learning} has an unoptimized implementation).
Although LDS boasts large improvements in raw accuracy at each scale, it also incurs a greater cost due to its expensive saliency generator.
Overall, the Pareto frontier for segmentation is very competitive, with ES dominating at $64\times64$, LDS at $128\times128$, and LZU at $256\times256$.
}
\label{fig:cost-performance}
\end{figure*}

\begin{table*}[t]
\centering
\footnotesize
\setlength{\tabcolsep}{2pt} 
\begin{tabular}{@{}rlccccccccccccccccccccc@{}}
\toprule
& & & \multicolumn{19}{c}{IOU} \\
\cmidrule(lr){4-22}
\multirow{-2}{*}{\raggedright\shortstack{Crop\\Size}} & Method & mIOU & road & swalk & build. & wall & fence & pole & tlight & sign & veg. & terr. & sky & person & rider & car & truck & bus & train & mbike & bike & \multirow{-2}{*}{\shortstack{Latency\\(ms)}} \\ \midrule
64 & Uniform & 26.4 & \textbf{93.9} & 35.6 & 68.6 & 3.5 & \textbf{2.9} & \textbf{0.5} & \textbf{0.0} & \textbf{0.1} & 72.5 & 21.1 & \textbf{76.0} & 26.8 & 0.9 & \textbf{57.1} & 8.9 & \textbf{16.9} & \textbf{8.0} & \textbf{0.0} & 8.2 & \textbf{15.5} \\
64 & LZU, fixed & \textbf{26.7} & 93.4 & \textbf{36.1} & \textbf{68.9} & \textbf{5.8} & 2.3 & 0.4 & \textbf{0.0} & 0.0 & \textbf{72.6} & \textbf{23.4} & 75.9 & \textbf{29.4} & \textbf{1.2} & 56.7 & \textbf{15.0} & 10.2 & 4.1 & \textbf{0.0} & \textbf{11.7} & 16.9 \\ \midrule
128 & Uniform & 39.3 & 96.3 & 54.0 & 78.4 & \textbf{15.0} & 7.9 & \textbf{8.1} & 8.5 & 16.6 & 81.2 & 34.4 & \textbf{86.7} & 42.9 & 13.8 & 74.4 & \textbf{22.9} & 41.6 & 24.4 & 10.2 & 29.6 & \textbf{16.1} \\
128 & LZU, fixed & \textbf{41.7} & \textbf{96.4} & \textbf{55.2} & \textbf{78.7} & 12.7 & \textbf{13.4} & \textbf{8.1} & \textbf{11.4} & \textbf{19.0} & \textbf{81.7} & \textbf{39.0} & 86.5 & \textbf{45.7} & \textbf{17.9} & \textbf{76.8} & 21.9 & \textbf{48.2} & \textbf{31.7} & \textbf{11.6} & \textbf{36.3} & 18.0 \\ \midrule
256 & Uniform & 53.6 & 97.5 & 64.0 & 84.7 & 20.0 & 19.0 & \textbf{22.1} & 34.8 & 41.6 & \textbf{87.0} & 41.9 & \textbf{91.2} & 59.3 & 33.7 & 84.1 & 39.2 & 62.9 & \textbf{57.9} & 27.7 & 49.1 & \textbf{19.1} \\
256 & LZU, fixed & \textbf{55.1} & \textbf{97.7} & \textbf{67.0} & \textbf{84.9} & \textbf{24.4} & \textbf{24.4} & 21.3 & \textbf{35.2} & \textbf{42.9} & \textbf{87.0} & \textbf{44.5} & 90.7 & \textbf{61.5} & \textbf{35.7} & \textbf{85.7} & \textbf{40.8} & \textbf{67.9} & 52.8 & \textbf{29.3} & \textbf{53.4} & 21.2 \\ \midrule
512 & Uniform & 63.8 & \textbf{98.3} & 73.3 & \textbf{88.8} & 29.2 & 34.3 & \textbf{40.6} & 54.4 & 61.6 & \textbf{90.7} & \textbf{47.7} & \textbf{94.0} & \textbf{72.7} & \textbf{50.6} & 89.1 & 45.6 & 72.1 & 59.1 & 44.5 & 64.9 & \textbf{32.3} \\
512 & LZU, fixed & \textbf{64.2} & \textbf{98.3} & \textbf{73.4} & 88.6 & \textbf{30.0} & \textbf{35.7} & 38.8 & \textbf{56.0} & \textbf{63.8} & 90.4 & 47.0 & 93.4 & 72.4 & 43.9 & \textbf{90.1} & \textbf{50.5} & \textbf{76.4} & \textbf{59.6} & \textbf{45.4} & \textbf{65.3} & 34.4 \\ \bottomrule
\end{tabular}
\caption{Full semantic segmentation results of PSPNet~\cite{zhao2017pyramid} on Cityscapes~\cite{cordts2016cityscapes}. At each resolution, LZU outperforms uniform downsampling.
}
\label{tab:seg-full}
\end{table*}

Results are given in Table~\ref{tab:det-avhd}. We outperform both uniform downsampling and FOVEA in all but one case, while incurring an additional latency of less than $4$ms.
The one exception is adaptive LZU at $0.75$x, which is evidence that our adaptive saliency hyperparameters, chosen at $0.5$x scale, struggle to generalize to other resolutions.
We also confirm that using cascaded saliency to train adaptive LZU is crucial. Although adaptive LZU outperforms fixed LZU at $0.5$x scale, plotting the accuracy-latency curves (Figure~\ref{fig:cost-performance}) reveals that fixed LZU is Pareto optimal at all points.

Finally, we explore how LZU performs in the \textit{upsampling} regime. We reuse the same models trained in our previous experiments, testing them with different pre-resampling resolutions. Results are shown in Table~\ref{tab:upsampling}. In this regime, LZU consistently outperforms uniform downsampling, even though information retention is no longer a factor.

\subsection{Semantic Segmentation}

For our semantic segmentation experiments, we compare to previous works ES~\cite{marin2019efficient} and LDS~\cite{jin2021learning}, so we adopt their setup.
We test the PSPNet~\cite{zhao2017pyramid} model (with a ResNet-50 backbone~\cite{he2016deep} and FPN~\cite{lin2017feature}) on Cityscapes~\cite{cordts2016cityscapes}. Cityscapes is an urban scene dataset with high resolution $1024 \times 2048$ images and $19$ classes. We perform our experiments at several image scales ($64\times 64$, $128 \times 128$, $256\times 256$, and $512\times 512$), taken by resizing a centered square crop of the input image. Our simple baseline trains and tests PSPNet with uniform downsampling. To reduce overfitting, we allot 500 images from the official training set into a mini-validation split.
We train our model on the remaining training images and evaluate at 10 equally spaced intervals on the mini-validation split.
We choose the best performing model and evaluate it on the official validation set.

For our LZU model, we unzoom spatial features after the FPN and use a fixed saliency map. Inspired by the idea of zooming on semantic boundaries~\cite{marin2019efficient}, we generate our fixed saliency by averaging the ground truth semantic boundaries over the train set. Notably, our saliency hyperparameters are chosen qualitatively (for producing a reasonably strong warp) and tested one-shot.

We report our full results in Table~\ref{tab:seg-full} and compare to previous works in Table~\ref{tab:seg}. Since our baseline results are slightly different than reported in previous works~\cite{marin2019efficient,jin2021learning}, we compare results using a percent change relative to the corresponding baseline. We find increased performance over the baseline at all scales, and at $256\times 256$, we beat both previous works with only $2.3$ms of additional latency. Plotting the accuracy-FLOPs tradeoff (Figure~\ref{fig:cost-performance}) reveals that the large improvements of LDS~\cite{jin2021learning} at $64\times64$ and $128\times128$ input scales come at significant cost in FLOPs. In actuality, ES~\cite{marin2019efficient} is Pareto optimal at $64\times64$ and $128\times128$, LDS~\cite{jin2021learning} at $128\times128$, and LZU at $256\times256$. We hypothesize that further improvements might be possible using an adaptive, learned formulation for saliency.

\subsection{Monocular 3D Object Detection}

\begin{table}[t]
\centering
\footnotesize
\begin{tabular}{@{}cr@{~}lr@{~}lr@{~}l@{}}
\toprule
& \multicolumn{6}{c}{Downsampled Resolution}\\
\cmidrule{2-7}
\rule{0pt}{1.1em} Method & \multicolumn{2}{c}{$64 \times 64$} & \multicolumn{2}{c}{$128\times128$} & \multicolumn{2}{c}{$256\times256$} \\
\midrule
Uniform (theirs) & \multicolumn{2}{c}{29} & \multicolumn{2}{c}{40} & \multicolumn{2}{c}{54} \\
Uniform (ours) & \multicolumn{2}{c}{26.4} &	\multicolumn{2}{c}{39.3} &	\multicolumn{2}{c}{53.6} \\
\midrule
ES~\cite{marin2019efficient} & 32 &(+10.3\%) & 43 &(+7.5\%) & 54 &(+0.0\%) \\
LDS~\cite{jin2021learning} & 36 &(\textbf{+24.1\%}) & 47& (\textbf{+17.5\%}) & 55 &(+1.9\%) \\
LZU, fixed & 
26.7 &(+1.1\%) &
41.7& (+6.1\%) &
55.1& (\textbf{+2.9\%}) \\
\bottomrule
\end{tabular}
\caption{Semantic segmentation results of PSPNet~\cite{zhao2017pyramid} on Cityscapes~\cite{cordts2016cityscapes}, in mIOU. Due to differing implementation, the performance of our baseline varies from reported values, so we report relative improvements. At $256\times 256$, we outperform prior works. At $64 \times 64$ and $128\times 128$, LZU performs worse than prior work, perhaps because "unzooming" features at such small scales is more destructive. We posit the performance losses from such aggressive downsampling factors (across all methods) may be too impractical for deployment, and so focus on the $256\times256$ regime.
}
\label{tab:seg}
\end{table}

\begin{table}[t]
\centering
\resizebox{\columnwidth}{!}{%
\setlength{\tabcolsep}{2pt} 
\footnotesize
\begin{tabular}{rlcccccccc}
\toprule
Scale & Method & NDS & mAP & mATE & mASE & mAOE  & mAVE & mAAE & Lat (ms) \\
\midrule
0.25x & Uniform & 21.8&	11.4&	\textbf{96.7}&	32.6&	90.1&	\textbf{125.0}&	\textbf{19.8} & \textbf{54.7} \\
0.25x & LZU, fixed & \textbf{23.4}&	\textbf{13.1}&	96.8&	\textbf{31.9}&	\textbf{82.7}&	129.4&	20.0 & 55.1 \\
\midrule
0.5x & Uniform & 27.5 &	17.5&	90.1&	28.8&	75.5&	131.6&	17.8 & \textbf{58.1} \\
0.5x & LZU, fixed & \textbf{29.3}	&\textbf{20.1}&	\textbf{88.9}&	\textbf{28.3}&	\textbf{73.9}	&\textbf{130.6}&	\textbf{16.7}& 58.5 \\
\midrule
0.75x & Uniform & 30.5&	21.0&	87.3&	27.9&	\textbf{67.0}&	\textbf{132.8}&	17.5 & \textbf{59.2} \\
0.75x & LZU, fixed & \textbf{31.8}&	\textbf{22.4}&	\textbf{83.8}&	\textbf{27.5}&	67.2&	134.6&	\textbf{15.9} & 59.7 \\
\midrule
1x & Uniform &31.2&	22.4&	\textbf{84.2}&	\textbf{27.4}&	70.9&	\textbf{129.6}&	\textbf{17.4} & \textbf{88.7} \\
1x & LZU, fixed &\textbf{32.6}&	\textbf{24.8}&	84.6&	27.5&	\textbf{68.2}&	131.6&	18.3& 89.4 \\
\bottomrule
\end{tabular}
}
\caption{
3D object detection results of FCOS3D~\cite{wang2021fcos3d} on nuScenes~\cite{nuScenes}.
Higher NDS and mAP is better, and lower is better on all other metrics.
Intuitively, size is an important cue for depth, and image deformations would stifle this signal. Suprisingly, this is \textit{not} the case.
LZU improves upon the uniform downsampling baseline at all scales with less than $1$ms of additional latency. Notably, LZU at $0.75$x scale even outperforms uniform downsampling at $1$x.
}
\label{tab:3ddet}
\end{table}


Finally, we evaluate LZU on monocular 3D object detection. To the best of our knowledge, no previous work has applied LZ downsampling to this task. The closest existing solution, FOVEA~\cite{thavamani2021fovea}, cannot be extended to 3D object detection, because 3D bounding boxes are amodal and cannot be unwarped in the same manner as 2D bounding boxes. For our base model, we use FCOS3D~\cite{wang2021fcos3d}, a fully convolutional model, with a ResNet-50 backbone~\cite{he2016deep} and FPN~\cite{lin2017feature}. For our dataset, we use nuScenes~\cite{nuScenes}, an autonomous driving dataset with multi-view $1600 \times 900$ RGB images for 1000 scenes and 3D bounding box annotations for 10 object classes. As is standard practice, we use the nuScenes Detection Score (NDS) metric, which is a combination of the usual mAP and measures of translation error (mATE), scale error (mASE), orientation error (mAOE), velocity error (mAVE), and attribute error (mAAE). We run experiments at $0.25$x, $0.5$x, $0.75$x, and $1$x scales and test against a uniform downsampling baseline. We train for 12 epochs with a batch size of 16 with default parameters as in MMDetection3D~\cite{mmdet3d2020}.

For our LZU model, again we unzoom post-FPN features and use a fixed saliency map. Inspired by FOVEA~\cite{thavamani2021fovea}, our fixed saliency is generated by using kernel density estimation on the set of projected bounding boxes in the image space. We reuse the same saliency hyperparameters from 2D detection. All other training settings are identical to the baseline.

Results are given in Table~\ref{tab:3ddet}. LZU performs consistently better than uniform downsampling, with less than $1$ms of additional latency.
Specifically, LZU improves mAP and the aggregate metric NDS, with mixed results on mATE, mASE, mAOE, mAVE, and mAAE.
Since the latter five metrics are computed on only \textit{true positives}, this demonstrates that LZU increases overall recall, while maintaining about equal performance on true positives.
Plotting the accuracy-latency curves (Figure~\ref{fig:cost-performance}) shows that LZU is Pareto optimal. We also repeat the same upsampling experiments as performed in 2D object detection. Results, shown in Table~\ref{tab:upsampling}, reaffirm the viability of LZU in the upsampling regime.

\section{Conclusion}
\label{sec:conclusion}

We propose LZU, a simple attentional framework consisting of "zooming" in on the input image, computing spatial features, and "unzooming" to invert any deformations. To unzoom, we approximate the forward warp as a piecewise bilinear mapping and invert each piece. LZU is highly general and can be applied to any task with 2D spatial input and any model with 2D spatial features. We demonstrate the versatility of LZU empirically on a variety of tasks and datasets, including monocular 3D detection which has never been done before. We also show that LZU may even be used when high-resolution sensor data is unavailable. For future work, we can consider alternatives to the "unzoom" formulation that are perhaps less destructive than simple resampling of features.

{\bf Broader impact.} Our work focuses on increasing the efficiency and accuracy of flagship vision tasks (detection, segmentation, 3D understanding) with high-resolution imagery. We share the same potential harms of the underlying tasks, but our approach may increase privacy concerns as identifiable information may be easier to decode at higher resolutions (e.g., facial identities or license plates). Because our approach is agnostic to the underlying model, it is reproducible with minimal changes to existing codebases.

\bigskip

\noindent {\bf Acknowledgements:} This work was supported by the CMU Argo AI Center for Autonomous Vehicle Research.
    \newpage
    {\small
    \bibliographystyle{ieee_fullname}
    \bibliography{egbib}
    }
    \clearpage
    \appendix

\section{Appendix}

\subsection{Bilinear Transformations}
\label{sec:bilinear-maps}

\begin{figure}[h]
\centering
\includegraphics[width=\columnwidth]{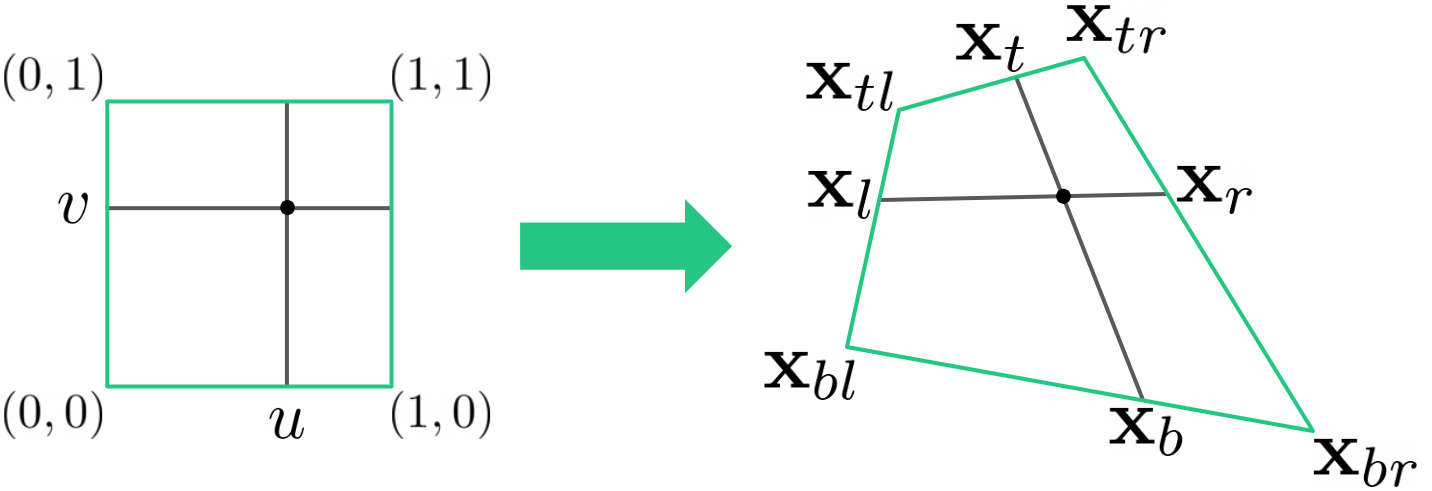}
\caption{
Geometric interpretation of bilinear transformations. Suppose we have such a transformation from the unit square to an arbitrary quadrilateral. Given coordinates $(u, v)$ in the unit square, if we draw lines in the quadrilateral such that
$u = \frac{|\bx_b \bx_{bl}|}{|\bx_{br} \bx_{bl}|} = \frac{|\bx_t \bx_{tl}|}{|\bx_{tr} \bx_{tl}|}$ and
$v = \frac{|\bx_l \bx_{bl}|}{|\bx_{tl} \bx_{bl}|} = \frac{|\bx_r \bx_{br}|}{|\bx_{tr} \bx_{br}|}$,
they will intersect at $\mathrm{BilinearTransformation}(u, v)$.
}
\label{fig:bilinear-map}
\end{figure}

Our construction in Section~\ref{sec:warp-inv} assumes prior knowledge of bilinear transformations. Bilinear transformations have actually been widely studied in the context of computer graphics~\cite{wolberg1990digital}. Here, for unfamiliar readers, we outline its definition and inverse formulation.

For simplicity, consider a bilinear transformation that maps the unit square to the quadrilateral with corners $\bx_{bl}, \bx_{br}, \bx_{tl}, \bx_{tr}$. True to its name, the transformation is defined \textit{within} the square via bilinear interpolation:
\begin{align}
    \mathrm{BilinearTransformation}(u, v) &= \bx_{bl} + (\bx_{br} - \bx_{bl})u \nonumber \\
    + (\bx_{tl} - \bx_{bl})v + (\bx_{tr} -& \bx_{br} - \bx_{tl} + \bx_{bl}) uv.
\end{align}
Interestingly, this transformation also has a geometric interpretation, shown in Figure~\ref{fig:bilinear-map}.

Now, consider the inverse of this mapping. Given a point $(x, y)$ in the quadrilateral, we want to find the point $(u, v)$ in the unit square that maps to it. A full derivation is given in~\cite{wolberg1990digital}, but if we define the following scalars
\begin{align}
    (a_0, b_0) &= \bx_{bl} \\
    (a_1, b_1) &= \bx_{br} - \bx_{bl} \\
    (a_2, b_2) &= \bx_{tl} - \bx_{bl} \\
    (a_3, b_3) &= \bx_{tr} - \bx_{br} - \bx_{tl} + \bx_{bl} \\
    c_0 &= a_1 (b_0 - y) + b_1 (x - a_0) \\
    c_1 &= a_3 (b_0 - y) + b_3 (x - a_0) + a_2 b_1 - a_2 b_1 \\
    c_2 &= a_3 b_2 - a_2 b_3,
\end{align}
then the solution $(u, v)$ must satisfy
\begin{equation}
\label{eq:bilinear-sol-1}
    c_2v^2 + c_1v + c_0 = 0
\end{equation}
and
\begin{equation}
\label{eq:bilinear-sol-2}
    u = \frac{x - a_0 - a_2 v}{a_1 + a_3 v}.
\end{equation}

Applying the quadratic formula on Equation~\ref{eq:bilinear-sol-1}, we can solve for $v$. Then, we can substitute into Equation~\ref{eq:bilinear-sol-2} to find $u$. Given a point $(x, y)$ in the quadrilateral, this will produce exactly one pair of solutions $(u, v)$ in the unit square (there may be extraneous solutions with $u$ or $v$ negative or greater than $1$).

Although these results assume a mapping from the unit square, they extend naturally to our use case. Recall from Section~\ref{sec:warp-inv} that $\mathrm{\widetilde\calT}_{ij}$ is a bilinear transformation from rectangle $R_{ij}$ to quadrilateral $\calT[R_{ij}]$. We can apply all previous results, simply by normalizing the coordinates within $R_{ij}$.

\subsection{Efficient Inversion of Nonseparable Warps}
\label{sec:nonsep-inversion}

\begin{figure*}[t]
\centering
\includegraphics[width=1.8\columnwidth]{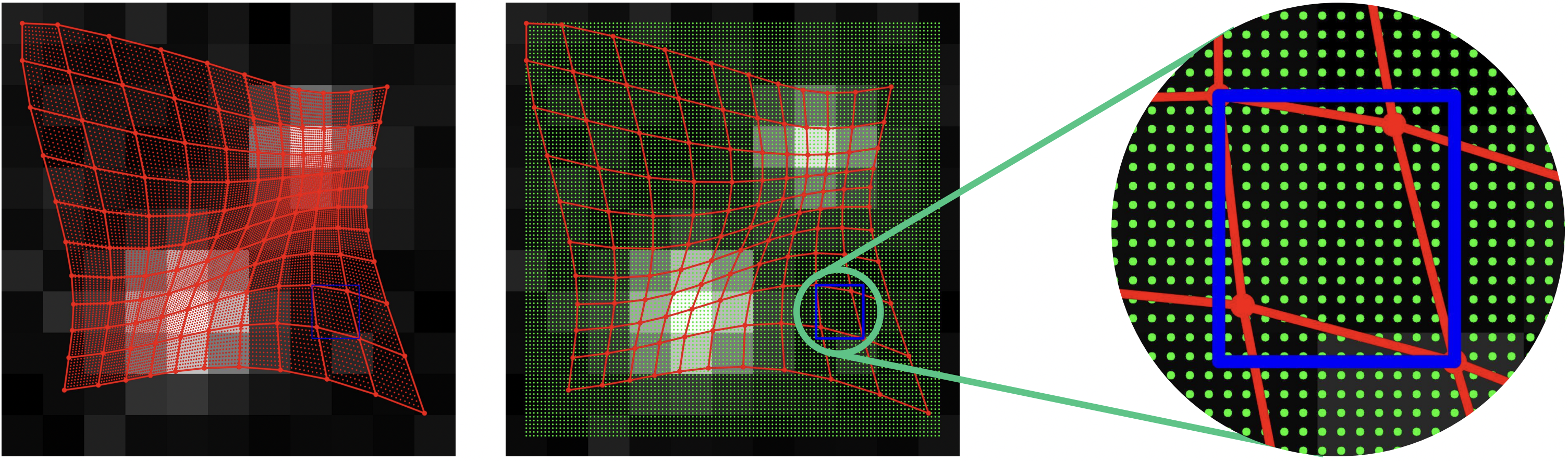}
\caption{
Unzooming in the nonseparable case. $\calT_{\mathrm{LZ}}$ is approximated as $\widetilde\calT_{\mathrm{LZ}}$, which is a $(h-1) \times (w-1)$ tiling of bilinear transformations (left).
We wish to compute $\widetilde\calT_{\mathrm{LZ}}^{-1}(\bx)$ at each green point $\bx \in \mathrm{Grid}(H'', W'')$, where $H'' \times W''$ is the desired output size (middle).
For the $ij$-th tile, we consider the set of all candidate green point $\bx$ within the enclosing blue box (right) and apply the corresponding inverse bilinear transformation. We set $\widetilde\calT_{\mathrm{LZ}}^{-1}(\bx) = \mathrm{BilinearTransformation}_{ij}^{-1}(\bx)$ only if it falls in the $ij$-th grid rectangle $R(i,j)$.
}
\label{fig:unzoom-nonsep}
\end{figure*}

In Section~\ref{sec:lzu}, we detail how to efficiently invert separable zooms $\calT_{\mathrm{LZ,sep}}$. To invert nonseparable zooms $\calT_{\mathrm{LZ}}$, 
it no longer suffices to invert each axis. We must instead reason in the full 2D space.

Suppose we have a nonseparable zoom $\calT_{\mathrm{LZ}}$. We compute $\calT_{\mathrm{LZ}}[\mathrm{Grid}(h, w)]$ for small $h, w$ and use this to approximate the forward zoom as $\widetilde\calT_{\mathrm{LZ}}$, an $(h-1) \times (w-1)$ piecewise tiling of bilinear maps. Now, to unzoom to a desired output resolution of $H'' \times W''$, we must evaluate $\widetilde\calT_{\mathrm{LZ}}[\mathrm{Grid}(H'', W'')]$. That is, for each $\bx \in \mathrm{Grid}(H'', W'')$, we must determine which of the $(h-1)(w-1)$ quadrilateral pieces it falls in and apply the corresponding inverse bilinear map. Recall from Appendix~\ref{sec:bilinear-maps} that applying an inverse bilinear map amounts to solving a quadratic.

In our actual implementation, we parallelize operations as much as possible. For the $ij$-th tile, instead of first determining which points $\bx \in \mathrm{Grid}(H'', W'')$ are inside of it and then applying the $ij$-th inverse bilinear map, we implement it the other way around. We consider a set of candidate interior points, apply the $ij$-th inverse bilinear map to all of them, and keep only those with a valid solution. The candidate points are those falling inside the axis-aligned rectangle enclosing that tile. The full procedure is described in Algorithm~\ref{alg:unzoom} and visualized in Figure~\ref{fig:unzoom-nonsep}. 

Our implementation takes about $12.6$ms to invert a nonseparable warp with $(h, w) = (31, 51)$ and an output shape $(H'', W'')=(600, 960)$, as in our Argoverse-HD~\cite{li2020towards} experiments. While this is not fast enough to support favorable accuracy-latency tradeoffs, we believe that further optimization (\eg using custom CUDA operations) may change this.

\begin{spacing}{1.2}
\begin{algorithm}
  \footnotesize
  \caption{Inverting nonseparable zooms $\calT_{\mathrm{LZ}}$. \\{\footnotesize In practice, we make the following optimizations. We vectorize the loop on line 13. We also fix $B_{ij}$ to be the max size over choices of $(i,j)$, allowing us to implement the loop on line 4 using batch-processing.}}
  \label{alg:unzoom}
  \setstretch{1.35}
    \begin{algorithmic}[1]
      \LineComment{See Appendix~\ref{sec:nonsep-inversion} for the algorithm setup and meaning of variables.}
      \Function{Unzoom}{$\calT_{\mathrm{LZ}}[\mathrm{Grid}(h, w)], (H'', W'')$}
        \State Initialize $\calT_{\mathrm{LZ}}^{-1}(\bx) = (0, 0)$ for all $\bx \in \mathrm{Grid}(H'', W'')$
        \For{$(i, j) \in [h-1] \times [w-1]$}
        
            \LineComment{corners of $R(i,j)$}
            \State $\bx_{tl}', \bx_{tr}' = 
            \left( \frac{i}{h-1}, \frac{j-1}{w-1} \right),
            \left( \frac{i}{h-1}, \frac{j}{w-1} \right)$
            \State $\bx_{bl}', \bx_{br}' = 
            \left( \frac{i}{h-1}, \frac{j-1}{w-1} \right),
            \left( \frac{i}{h-1}, \frac{j}{w-1} \right)$
            \LineComment{corners of $ij$-th quadrilateral tile}
            \State $\bx_{tl}, \bx_{tr} = \calT_{\mathrm{LZ}}^{-1}\left(\bx_{tl}' \right),
            \calT_{\mathrm{LZ}}^{-1}\left(\bx_{tr}' \right)$
            \State $\bx_{bl}, \bx_{br} =
            \calT_{\mathrm{LZ}}^{-1}\left(\bx_{bl}' \right),
            \calT_{\mathrm{LZ}}^{-1}\left(\bx_{br}' \right)$
            \LineComment{top-left and bottom-right corners of rectangle}
            \LineComment{enclosing the quadrilateral tile}
            \State $\bc_{tl} = \min(\bx_{tl}, \bx_{tr}, \bx_{bl}, \bx_{br})$
            \State $\bc_{br} = \max(\bx_{tl}, \bx_{tr}, \bx_{bl}, \bx_{br})$
            \LineComment{set of all candidate points in the $ij$-th tile}
            \State $B_{ij} = \left\{ \bx \in \mathrm{Grid}(H'', W'') : \bc_{tl} \leq \bx \leq \bc_{br} \right\}$
            \For{$\bx \in B_{ij}$}
                \State $\bx' = \mathrm{BilinearTransformation}^{-1}_{ij}(\bx)$
                \If{$\bx_{tl}' \leq \bx' \leq \bx_{br}'$}
                \State $\calT_{\mathrm{LZ}}^{-1}(\bx) = \bx'$
                \EndIf
            \EndFor
        \EndFor
        \State \Return $\calT^{-1}_{\mathrm{LZ}}$
      \EndFunction

    \end{algorithmic}
\end{algorithm}
\end{spacing}

\subsection{Analysis of our Warping Approximations}
\label{sec:approximations}

We make two approximations in our formulation.
First, to ensure that the composition of forward and inverse warps is truly the identity function, we use the approximate forward warp $\widetilde\calT$ in place of the true forward warp $\calT$.
This trades latency for how well $\widetilde\calT$ zooms in on the intended regions of interest.
See Figure~\ref{fig:warp_approx} for a visualization of this effect.

\begin{figure*}
    \centering
    \includegraphics[width=\linewidth]{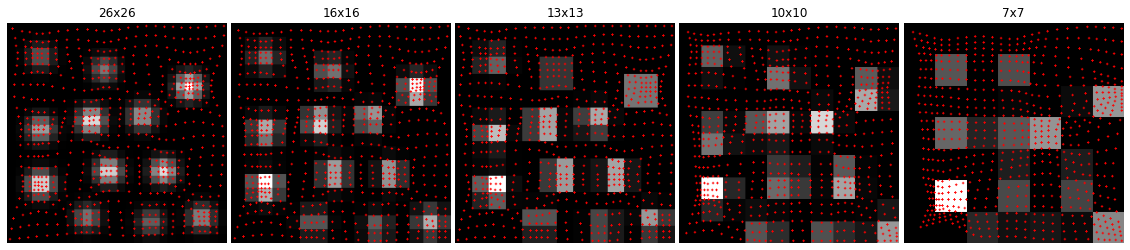}
    \caption{Quality of the approximate forward warp $\widetilde\calT$ as we decrease the dimensions $h \times w$ of our piecewise bilinear approximation. 
    $\widetilde\calT$ degrades noticeably once $h, w$ decrease beyond a certain threshold.
    }
    \label{fig:warp_approx}
\end{figure*}

Second, we use bilinear downsampling to approximate the inverse at lower resolutions after feature pyramid networks~\cite{lin2017feature}.
This approximation is surprisingly effective!
For our fixed LZU model on 2D detection, we calculate that the error between $\widetilde\calT^{-1}$ and the doubly approximated inverse $\widehat\calT_d^{-1}$ given by
\begin{equation}
    \underset{\bx\in\mathrm{Grid}(H'/d, W'/d)}{\mathrm{avg}} \left\|\widetilde\calT^{-1}(\bx) - \widehat\calT_d^{-1}(\bx)\right\|_2
\end{equation}
is only $0.0274$, $0.0065$, $0.0028$ pixels at $d=2$, $4$, and $8$!

\subsection{Sensitivity to Saliency}
\label{sec:sens-saliency}

Fixed LZU is quite robust to choice of saliency.
Details on how we computed our saliency maps are given in Appendix~\ref{sec:impl-details}.
For 2D detection, we performed a grid search at $0.5$x scale to determine the saliency hyperparameters.
The results show that LZU is quite robust to choice of saliency, as long as the warp is not too strong (see Table~\ref{tab:saliency_sensitivity}).
For semantic segmentation and 3D detection, we chose saliency one-shot, which suggests that these fixed LZU models are also robust to choice of saliency.

Our grid search suggests that adaptive LZU is also robust to choice of hyperparameters than fixed LZU.
However, these saliency hyperparameters, chosen at $0.5$x scale, struggle to generalize to $0.75$x and $1$x scale (see Table~\ref{tab:det-avhd}). As a result, for adaptive warps, it may be necessary to tune saliency at each scale.

\begin{table}[t]
\centering
\setlength{\tabcolsep}{3pt} 
\footnotesize
\centering
\vspace{0.5em}
\begin{minipage}{.5\linewidth}
\centering
LZU, fixed
\begin{tabular}{rccccc}
    \toprule
    & \multicolumn{4}{c}{Attraction Kernel fwhm} \\
    \cmidrule(r){2-5}
    Amp.~~ & 4 & 10 & 16 & 22 \\
    \cmidrule(r){1-1}\cmidrule(r){2-5}
    1&\bf\textcolor[rgb]{0.479087719298246, 0.702951496388029, 0.54205572755418}{34.6}&\bf\textcolor[rgb]{0.446789473684211, 0.685386996904025, 0.51761919504644}{34.8}&\bf\textcolor[rgb]{0.349894736842106, 0.632693498452013, 0.44430959752322}{35.4}&\bf\textcolor[rgb]{0.301447368421052, 0.606346749226006, 0.40765479876161}{35.7}\\
    5&\bf\textcolor[rgb]{0.495236842105263, 0.711733746130031, 0.55427399380805}{34.5}&\bf\textcolor[rgb]{0.462938596491228, 0.694169246646027, 0.529837461300309}{34.7}&\bf\textcolor[rgb]{0.802070175438597, 0.87859649122807, 0.786421052631579}{32.6}&\bf\textcolor[rgb]{0.479087719298246, 0.702951496388029, 0.54205572755418}{34.6}\\
    10&\bf\textcolor[rgb]{0.446789473684211, 0.685386996904025, 0.51761919504644}{34.8}&\bf\textcolor[rgb]{0.511385964912281, 0.720515995872033, 0.56649226006192}{34.4}&\bf\textcolor[rgb]{0.85051754385965, 0.904943240454077, 0.82307585139319}{32.3}&\bf\textcolor[rgb]{0.495236842105263, 0.711733746130031, 0.55427399380805}{34.5}\\
    50&\bf\textcolor[rgb]{0.414491228070176, 0.667822497420021, 0.4931826625387}{35.0}&\bf\textcolor[rgb]{0.656728070175439, 0.799556243550052, 0.67645665634675}{33.5}&\bf\textcolor[rgb]{0.9	,0.58	,0.54}{31.4}&\bf\textcolor[rgb]{0.624429824561403, 0.781991744066047, 0.652020123839009}{33.7}\\
    100&\bf\textcolor[rgb]{0.414491228070176, 0.667822497420021, 0.4931826625387}{35.0}&\bf\textcolor[rgb]{0.608280701754387, 0.773209494324046, 0.63980185758514}{33.8}&\bf\textcolor[rgb]{0.9	,0.58,	0.54}{31.3}&\bf\textcolor[rgb]{0.705175438596491, 0.825902992776058, 0.713111455108359}{33.2}\\
    \bottomrule
\end{tabular}
\end{minipage}%
\begin{minipage}{.5\linewidth}
\centering
LZU, adaptive
\begin{tabular}{rccccc}
    \toprule
    & \multicolumn{4}{c}{Attraction Kernel fwhm} \\
    \cmidrule(r){2-5}
    Amp.~~ & 4 & 10 & 16 & 22 \\
    \cmidrule(r){1-1}\cmidrule(r){2-5}
    1&\bf\textcolor[rgb]{0.430640350877193, 0.676604747162023, 0.50540092879257}{34.9}&\bf\textcolor[rgb]{0.253, 0.58, 0.371}{36}&\bf\textcolor[rgb]{0.31759649122807, 0.615128998968008, 0.41987306501548}{35.6}&\bf\textcolor[rgb]{0.333745614035088, 0.62391124871001, 0.43209133126935}{35.5}\\
    5&\bf\textcolor[rgb]{0.527535087719299, 0.729298245614035, 0.57871052631579}{34.3}&\bf\textcolor[rgb]{0.398342105263158, 0.659040247678019, 0.48096439628483}{35.1}&\bf\textcolor[rgb]{0.689026315789475, 0.817120743034056, 0.70089318885449}{33.3}&\bf\textcolor[rgb]{0.398342105263158, 0.659040247678019, 0.48096439628483}{35.1}\\
    10&\bf\textcolor[rgb]{0.672877192982457, 0.808338493292054, 0.68867492260062}{33.4}&\bf\textcolor[rgb]{0.527535087719299, 0.729298245614035, 0.57871052631579}{34.3}&\bf\textcolor[rgb]{0.866666666666667, 0.913725490196078, 0.835294117647059}{32.2}&\bf\textcolor[rgb]{0.575982456140351, 0.755644994840042, 0.6153653250774}{34.0}\\
    50&\bf\textcolor[rgb]{0.608280701754387, 0.773209494324046, 0.63980185758514}{33.8}&\bf\textcolor[rgb]{0.705175438596491, 0.825902992776058, 0.713111455108359}{33.2}&\bf\textcolor[rgb]{0.85051754385965, 0.904943240454077, 0.82307585139319}{32.3}&\bf\textcolor[rgb]{0.543684210526316, 0.738080495356037, 0.590928792569659}{34.2}\\
    100&\bf\textcolor[rgb]{0.575982456140351, 0.755644994840042, 0.6153653250774}{34.0}&\bf\textcolor[rgb]{0.640578947368421, 0.79077399380805, 0.664238390092879}{33.6}&\bf\textcolor[rgb]{0.85051754385965, 0.904943240454077, 0.82307585139319}{32.3}&\bf\textcolor[rgb]{0.624429824561403, 0.781991744066047, 0.652020123839009}{33.7}\\
    \bottomrule
\end{tabular}
\end{minipage}
\caption{Grid search over hyperparameters for 2D object detection 
 (see Appendix~\ref{sec:impl-details}) shows that LZU is quite robust to choice of saliency. 
For comparison, uniform downsampling yields an AP of $32.2$.}
\label{tab:saliency_sensitivity}
\end{table}

\subsection{Additional Results}

\begin{figure}[t]
    \centering
    \includegraphics[width=\columnwidth]{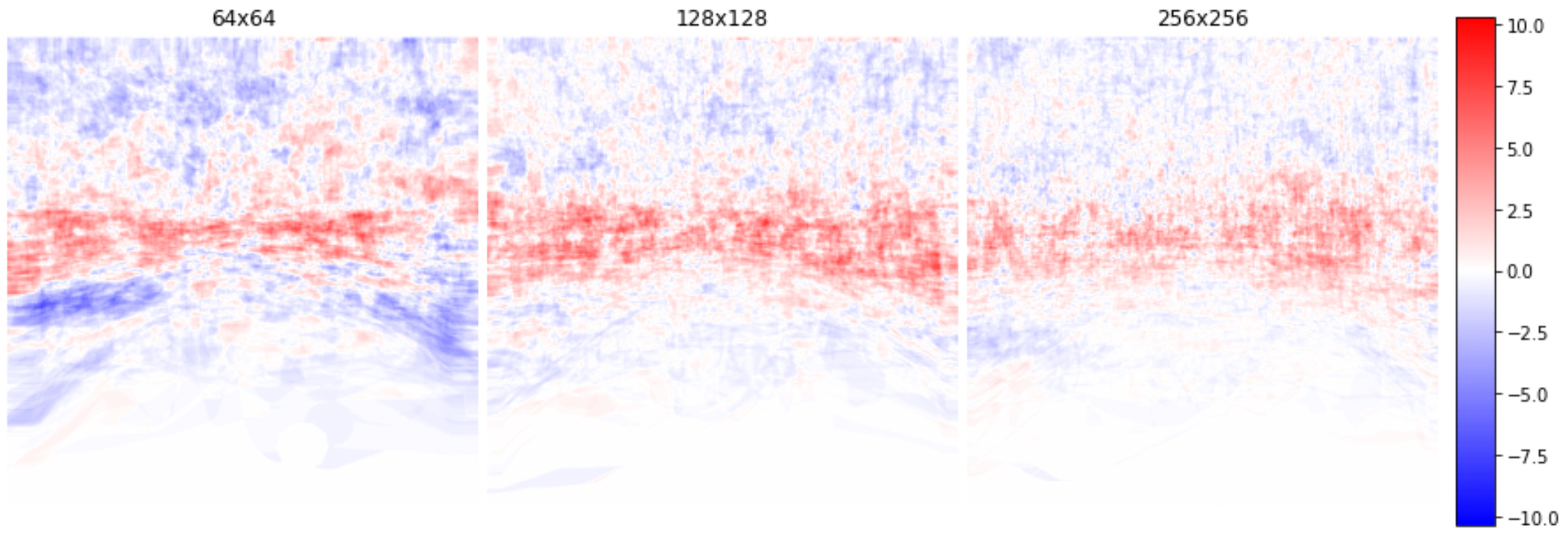}
    \caption{Visualizations of the per-pixel accuracy difference between the uniform and the LZU models at each resolution. The largest gains occur at the central horizontal strip, largely corresponding with the saliency map used for segmentation. However, performance also \textit{decreases} at the top and bottom of the image. Thus, using a fixed saliency map must be conscious decision to trade performance at low saliency regions for high saliency regions. We hypothesize that adaptive saliency maps could lead to more uniform gains.}
    \label{fig:spatial_accuracy}
\end{figure}

In Figure~\ref{fig:spatial_accuracy}, we plot the spatial accuracy of our CityScapes~\cite{cordts2016cityscapes} models.

\subsection{Implementation Details}
\label{sec:impl-details}

Our experiments are implemented using open-source libraries MMDetection~\cite{mmdetection}, MMSegmentation~\cite{mmseg2020}, and MMDetection3D~\cite{mmdet3d2020}, all released under the Apache 2.0 License.
We use GeForce RTX 2080 Ti's for training and training, which takes at most $5$ GPU-days for any given model, but the precise amount varies by model and task. We perform all timing experiments with a batch size of 1 on a single GPU.

\paragraph{Argoverse-HD~\cite{li2020towards} Detection}
As done in FOVEA~\cite{thavamani2021fovea}, for our uniform downsampling experiments, we finetune a COCO-pretrained model for 3 epochs with the random left-right image flip augmentation, a batch size of 8, momentum SGD with a learning rate of 0.005, momentum of 0.9, weight decay of 1e-4, learning rate warmup for 1000 iterations, and a per-iteration linear learning rate decay~\cite{li2019budgeted}.

For both LZU models, we use a learning rate of $0.01$ and keep all other hyperparameters identical to the baseline. To "zoom", we use a $31 \times 51$ saliency map and the separable anti-cropping formulation $\calT_{\mathrm{LZ,sep,ac}}$ (as proposed in~\cite{thavamani2021fovea} and discussed in Section~\ref{sec:saliency-downsampling}).

For the fixed saliency LZU model, we use Gaussian distance kernels $k_x$ and $k_y$ of full-width-half-maximum (fwhm) $22$. 
To generate the fixed saliency map, we use kernel density estimation (KDE) on all training bounding boxes with hyperparameters amplitude $a=1$ and bandwidth $b=64$. For details on the effects of $a$ and $b$, refer to~\cite{thavamani2021fovea}.

For the adaptive saliency LZU model, we use Gaussian distance kernels $k_x$ and $k_y$ of fwhm 10. To generate adaptive saliency, we use KDE on detections from the previous frame with $a=1$ and $b=64$. When training, to simulate motion, we jitter bounding boxes by $\calN(0, 7.5)$ pixels horizontally and $\calN(0, 3)$ pixels vertically.

For each LZU experiment, we run a grid search at $0.5$x scale over separable/nonseparable, amplitude $a=1,5,10,50,100$, and distance kernel's $\text{fwhm}=4,10,16,22$ to determine optimal settings. This is done using an $80/20$ split of the train set, so as to not overfit on the real validation set. We generate this split such that locations between splits are disjoint. All other hyperparameters are chosen one-shot.



\paragraph{Cityscapes~\cite{cordts2016cityscapes} Segmentation}
We train the uniform downsampling baseline using mostly the default hyperparameters from MMSegmentation~\cite{mmseg2020}.
The only changes are to the data augmentation pipeline and the evaluation frequency.
Comprehensively, we train with just the photometric distortion augmentation (random adjustments in brightness, contrast, saturation, and hue), a batch size of 16, momentum SGD with a learning rate of 0.01, momentum of 0.9, weight decay of 5e-4, and a polynomial learning rate schedule with power 0.9. To account for overfitting, we validate our performance at 10 equally spaced intervals on a 500-image subset of the training dataset (and train on the others). For all experiments, we train for 80K iterations, with the exception of $64\times64$, which we train for 20K iterations due to rapid overfitting.

For the fixed LZU model, we use the same training hyperparameters as the baseline. To "zoom", we use the separable anti-cropping formulation $\calT_{\mathrm{LZ,sep,ac}}$ with a $45 \times 45$ saliency map and Gaussian distance kernels $k_x$, $k_y$ of fwhm $15$. To generate the fixed saliency, we aggregate ground truth semantic boundaries over the train set. Precisely, we define boundaries to be pixels which differ from at least one of its eight neighbors. We compute semantic boundaries for each $256 \times 256$ ground truth segmentation, assign boundaries an intensity of $200$ and background an intensity of $1$, and average pool down to a $45 \times 45$ saliency map. The semantic boundary intensity value was chosen qualitatively (for producing a reasonably strong warp) and tested one-shot.
 
\paragraph{nuScenes~\cite{nuScenes} 3D Detection}
We train the uniform downsampling baseline using all default hyperparameters from MMDetection3D~\cite{mmdet3d2020}, except the learning rate, which we reduce for stability. 
Specifically, we train for 12 epochs with the random left-right flip augmentation, a batch size of 16, momentum SGD with a learning rate of 0.001, momentum of 0.9, weight decay of 1e-4, doubled learning rate on bias parameters with no weight decay, L2 gradient clipping, a step learning rate schedule with drops at epochs 8 and 11, and a linear learning rate warmup for the first 500 iterations.

For the fixed LZU model, we use the same training hyperparameters as the baseline. To "zoom", we use the separable anti-cropping formulation $\calT_{\mathrm{LZ,sep,ac}}$ with a $27 \times 48$ saliency map and Gaussian distance kernels $k_x$, $k_y$ of fwhm $10$. To generate the fixed saliency, we project 3D bounding boxes into the image plane and reuse the same KDE formulation with the same hyperparameters ($a=1$ and $b=64$) as used in 2D detection. These are all chosen and evaluated one-shot.



\fi

\end{document}